\documentclass[letterpaper]{article} 
\usepackage{aaai24}  
\usepackage{times}  
\usepackage{helvet}  
\usepackage{courier}  
\usepackage[hyphens]{url}  
\usepackage{graphicx} 
\usepackage{natbib}  
\usepackage{caption} 
\usepackage{algorithm}
\usepackage{algorithmic}

\usepackage{times}
\usepackage{epsfig}
\usepackage{graphicx}
\usepackage{amsmath}
\usepackage{amssymb}
\usepackage{url}
\usepackage{multirow}
\usepackage[table,xcdraw]{xcolor}
\usepackage{threeparttable}

\usepackage{bbding}
\usepackage{graphicx}
\usepackage{subfigure}
\usepackage{float}

\usepackage{booktabs}
\makeatletter
\renewcommand{\maketag@@@}[1]{\hbox{\m@th\normalsize\normalfont#1}}%
\makeatother
\usepackage[table,xcdraw]{xcolor}
\definecolor{mygray}{gray}{.9}
\usepackage{multirow}
\usepackage{soul}
\usepackage{color, xcolor}

\usepackage{newfloat}
\usepackage{listings}
\DeclareCaptionStyle{ruled}{labelfont=normalfont,labelsep=colon,strut=off} 
\floatstyle{ruled}
\newfloat{listing}{tb}{lst}{}
\floatname{listing}{Listing}
\title{Unifying Visual and Vision-Language Tracking via Contrastive Learning}
\author {
    Yinchao Ma\textsuperscript{\rm 1},
    Yuyang Tang\textsuperscript{\rm 1},
    Wenfei Yang\textsuperscript{\rm 1},
    Tianzhu Zhang\textsuperscript{\rm 1}\footnote{Corresponding author},
    Jinpeng Zhang\textsuperscript{\rm 2},
    Mengxue Kang\textsuperscript{\rm 2}
}
\affiliations {
    \textsuperscript{\rm 1}Deep Space Exploration Laboratory/School of Information Science and Technology, University of Science and Technology of China\\
    \textsuperscript{\rm 2}Intelligent Science Technology Academy of CASIC\\
    \{imyc,yuyangtang,yangwf\}@mail.ustc.edu.cn, tzzhang@ustc.edu.cn
}

\usepackage{bibentry}

\begin{document}

\maketitle

\begin{abstract}
Single object tracking aims to locate the target object in a video sequence according to the state specified by different modal references, including the initial bounding box (BBOX), natural language (NL), or both (NL+BBOX). 
Due to the gap between different modalities, most existing trackers are designed for single or partial of these reference settings and overspecialize on the specific modality.
Differently, we present a unified tracker called UVLTrack, which can simultaneously handle all three reference settings (BBOX, NL, NL+BBOX) with the same parameters.
The proposed UVLTrack enjoys several merits. First, we design a modality-unified feature extractor for joint visual and language feature learning and propose a multi-modal contrastive loss to align the visual and language features into a unified semantic space. 
Second, a modality-adaptive box head is proposed, which makes full use of the target reference to mine ever-changing scenario features dynamically from video contexts and distinguish the target in a contrastive way, enabling robust performance in different reference settings. 
Extensive experimental results demonstrate that UVLTrack achieves promising performance on seven visual tracking datasets, three vision-language tracking datasets, and three visual grounding datasets. 
Codes and models will be open-sourced at https://github.com/OpenSpaceAI/UVLTrack.
\end{abstract}

\section{Introduction}
Single object tracking is one of the fundamental research topics in computer vision, aiming to locate the target object in a video sequence according to the reference specified by the initial bounding box (BBOX)~\cite{vottask2}, natural language (NL)~\cite{nltrack}, or both (NL+BBOX)~\cite{TNL2K}.
It has a wide range of applications in robotics, video surveillance, autonomous driving, human-computer interaction and so on~\cite{vottask3}.
Although great progress has been achieved for specific reference settings, it is still challenging to design a unified tracker that performs well across all three reference settings.

\begin{figure}[t]
    \centering
    \includegraphics[width=\linewidth]{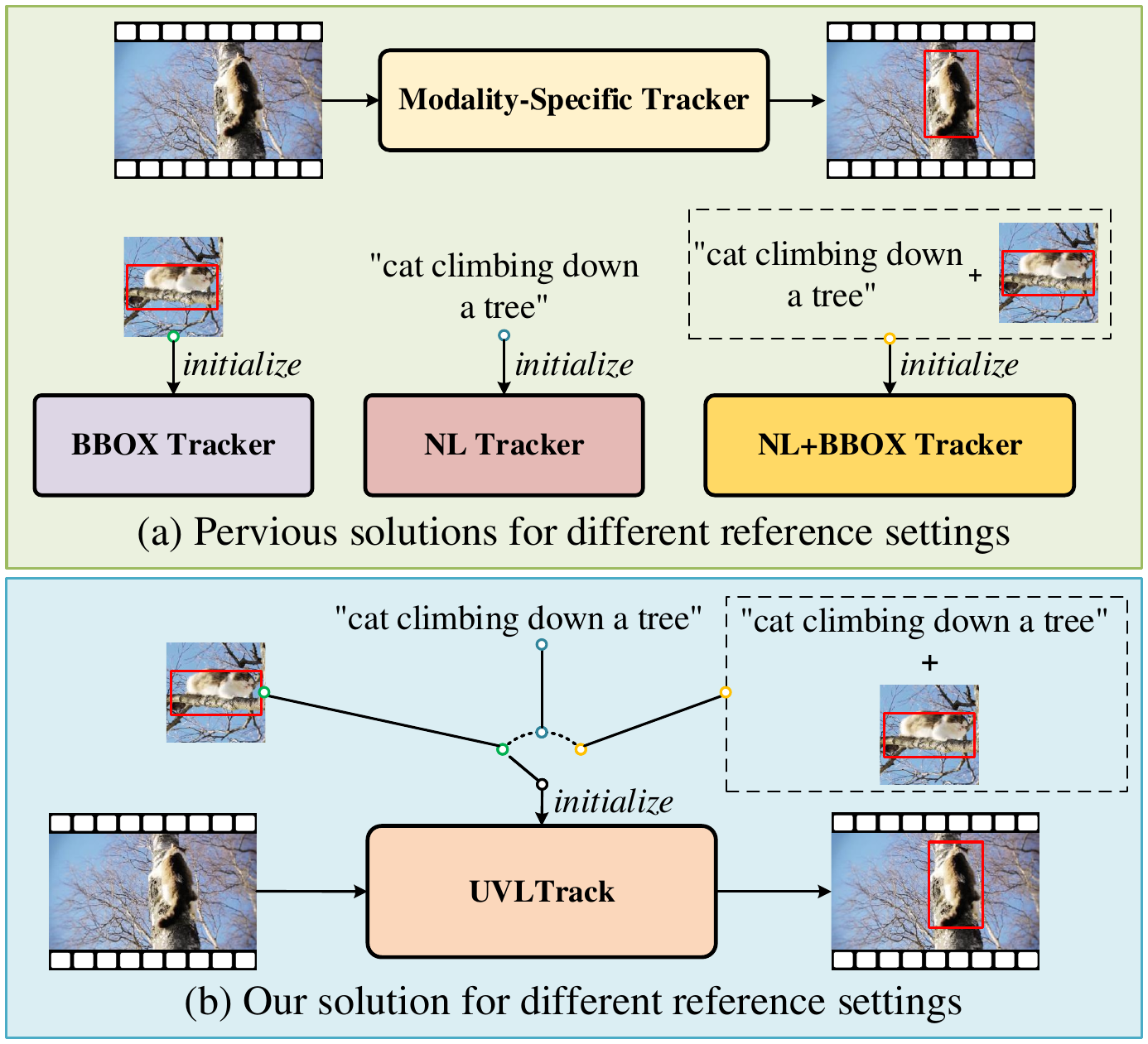}
    \caption{Comparison between previous solutions and UVLTrack. BBOX, NL, NL+BBOX tracker means the tracker is designed to utilize the bounding box, natural language, or both as the target reference respectively. Our UVLTrack can simultaneously handle three different reference settings.}
    \label{fig:introduction}
\end{figure}

Most trackers~\cite{SiamRPNplusplus,SiamFC++,yan2021learning} utilize the target bounding box in the first frame as the reference (\textbf{BBOX}).
They commonly crop a template according to the given bounding box and locate the target in subsequent frames by interacting with the cropped template.
Representatively, one-stream trackers~\cite{OSTrack,cui2022mixformer} combine feature extraction and interaction of the template and search region in Transformer architectures, achieving superior performance.
However, the bounding box has no direct target semantics, which may lead to ambiguity~\cite{TNL2K}.
Different from the above tracking paradigm, tracking by natural language specification~\cite{nltrack} provides a novel manner of human-computer interaction, which specifies the target based on the natural language reference (\textbf{NL}).
This task can be roughly divided into two steps. 1) locating the target in the first frame based on the language description. 2) tracking the target based on the language description and the predicted bounding box.
Recently, JointNLT~\cite{JointNLT} proposes a unified network to jointly conduct locating and tracking, enabling end-to-end model optimization.
For providing more accurate target reference, some trackers~\cite{VLT,SNLT} specify the target by both natural language and bounding box (\textbf{NL+BBOX}).
They embed language descriptions into visual features through dynamic filter~\cite{nltrack}, cross-correlation~\cite{SNLT} or channel-wise attention~\cite{VLT}, achieving more robust tracking.
Language description brings rich target semantics for tracking.
However, due to the semantic gap between different modalities, trackers designed with natural language show limited performance in the bounding box reference setting.

The modality of the target reference varies with the application scenario.
However, previous trackers overspecialize on the specific modalities, limiting their generalization, as shown in Figure~\ref{fig:introduction}.
To address the above limitation, we seek to combine visual and vision-language tracking into a unified framework.
It has two main benefits.
First, the unified tracker can simultaneously cope with three types of target references, enabling a wider range of application scenarios.
Second, we can utilize richer target references to optimize models, thereby improving their generalization ability.
By studying previous methods, we summarize two key issues that need to be considered to design the unified visual and vision-language tracking framework.

1) \textbf{Modality-aligned feature learning}.
Most vision-language trackers introduce natural language reference through well-designed fusion modules.
Typically, VLT~\cite{VLT} designs a ModaMixer to fuse the language feature and visual feature by channel-wise attention.
JointNLT~\cite{JointNLT} builds feature interactions between language and vision through self-attention mechanisms.
However, they ignore the semantic gap between different modalities, resulting in a tendency for trained visual-language trackers to rely on semantic information in language references, which limits their performance in pure bounding box reference setting~\cite{VLT}.
To this end, it is necessary to design a new modality-aligned feature extractor, achieving consistent feature learning for different modal references.
2) \textbf{Modality-adaptive target localization}.
Existing trackers design various static box heads to estimate the target state, such as anchor-free head~\cite{OSTrack}, corner-based head~\cite{JointNLT}, and point-based head~\cite{apmt}.
These heads commonly take the reference-enhanced features of the search region as input, and regress the target box through offline trained parameters.
However, various reference modalities increase the difficulty of static head training, which may lead to compromised results in different reference settings.
Thus, we argue that it is better to design a dynamic head, which can make full use of different modal references to mine ever-changing scenario features from video contexts to discriminate the target.

Motivated by the above discussions, we propose a unified framework for visual and vision-language tracking, termed UVLTrack, which mainly consists of a modality-unified feature extractor and a modality-adaptive box head.
The \textbf{modality-unified feature extractor} is constructed based on Transformer architecture, in which we extract features of different modalities separately in shallow encoder layers and fuse them in deep encoder layers.
Such a design avoids the confusion of low-level feature modeling between different modalities and allows high-level semantics interaction.
Besides, we design a multi-modal contrastive loss to align visual and language features into a unified semantic space, so as to realize consistent feature learning for different modal references.
The \textbf{modality-adaptive box head} dynamically mines ever-changing scenario information from video contexts and localize the target in a contrastive way.
Specifically, we propose a novel distribution-based cross-attention mechanism, which can make full use of different modal references to adaptively mine features of target, distractor and background from historic scenarios.
Then, the target can be localized directly through feature comparison.
By introducing dynamic scenario information, UVLTrack can achieve more robust tracking under different modal references.

To summarize, the main contributions of this work are:
(1) We propose a novel unified tracker, UVLTrack, for visual and vision-language tracking, which can simultaneously cope with three types of target reference (BBOX, NL, NL+BBOX).
(2) We design a modality-unified feature extractor for joint visual and language feature learning, and design a multi-modal contrastive loss to align different modal features into a unified semantic space.
(3) We propose a modality-adaptive box head to dynamically mine scenario features by different modal references and localize the target in a contrastive way, which helps UVLTrack achieve robust performance across all reference settings.
(4) Extensive experimental results on seven visual tracking datasets, three vision-language tracking datasets, and three visual grounding datasets demonstrate that UVLTrack shows promising performance compared with modality-specific counterparts.

\section{Related Work}
\subsection{Visual Tracking}
Visual tracking aims to locate the target in a video sequence according to the given bounding box in the first frame (BBOX).
Visual trackers commonly crop a target template from the first frame and estimate the target state in subsequent frames by interacting with the cropped template.
Siamese-based trackers~\cite{SiamRPNplusplus,SiamCAR} extract features from both template and search region with the siamese network and locate the target through well-designed matching modules.
Discriminative Correlation Filter based trackers~\cite{ATOM,DiMP} learn a correlation filter from historic target states in the video to discriminate the target from backgrounds.
Recently, one-stream trackers~\cite{cui2022mixformer,OSTrack} achieve joint feature extraction and interaction using Transformer architectures~\cite{wu2021cvt,2017Attention}, which simplify the tracking pipeline and achieve superior performance.
However, these visual trackers are designed purely based on visual features, which cannot flexibly introduce high-level language semantics to reduce visual ambiguity.

\begin{figure*}[t]
    \centering
    \includegraphics[width=1.0\linewidth]{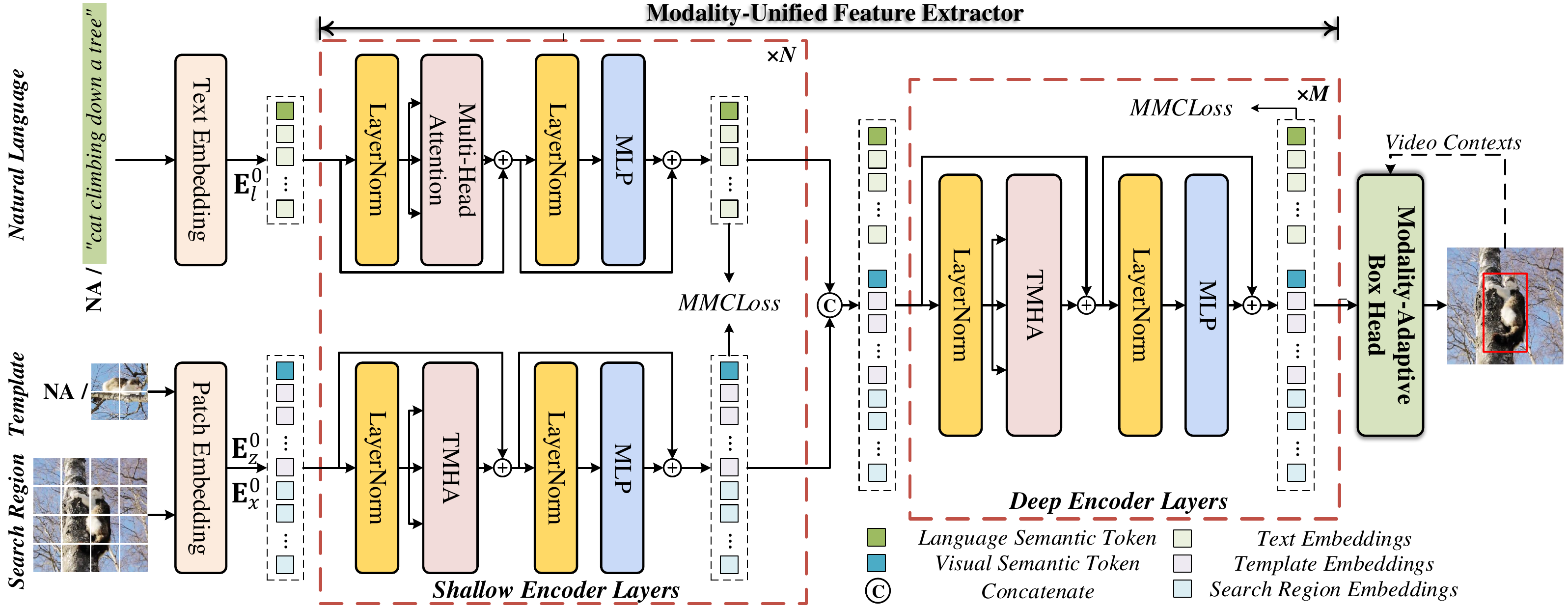}
    \caption{
    A unified tracking framework for different target references.
    {NA} means ``not available". Natural language is not available for visual tracking task and template is not available for grounding task.
    Different from previous trackers designed for specific reference modalities, our UVLTrack can simultaneously handle all target reference settings (BBOX, NL, NL+BBOX).
    }
    \label{fig:architecture}
\end{figure*}

\subsection{Vision-Language Tracking}
Natural language can provide clear target semantics to avoid visual ambiguity.
Thus, some trackers seek to utilize natural language to specify the tracking target.
\textbf{Tracking by natural language specification} (NL) provides a novel human-computer interaction manner, which specifies the target purely based on natural language.
Li \textit{et al.} first define this task and provides a baseline by combining a grounding model and a tracking model.
Then, some trackers~\cite{TSN,TNL2K,GTI} follow this paradigm to design different models to solve the grounding task and tracking task separately.
Recently, JointNLT~\cite{JointNLT} performs tracking and grounding using a unified model, which simplifies the overall framework and enables end-to-end optimization.
\textbf{Tracking by language and box specification} (NL+BBOX) specifies the target through both the initial bounding box and natural language.
Li \textit{et al.}~\cite{nltrack} firstly introduce natural language into tracking achieving more robust results than visual tracker, which demonstrates the potential of vision-language tracking.
SNLT~\cite{SNLT} embeds natural language into Siamese-based trackers as a convolutional kernel and locates the target through cross-correlation.
VLT~\cite{VLT} treats the natural language feature as a selector to weigh different visual feature channels, enhancing target-related channels for robust tracking.
Also, JointNLT~\cite{JointNLT} introduces natural language by interacting language and visual features in Transformer blocks.
However, due to the semantic gap between vision and language, these trackers trained with natural language show limited performance in pure bounding box reference setting~\cite{VLT}.
To this end, we design a multi-modal contrastive loss to align features of different modalities into a unified semantic space.
Meanwhile, a dynamic head, modality-adaptive box head, is proposed to alleviate the difficulty of static head training for different modal references.
Thanks to the effective designs, our UVLTrack achieves promising performance across all reference settings with high FPS.

\section{Method}
%
In this section, we first introduce the overall architecture of UVLTrack, which presents a simple but effective pipeline for unified visual and vision-language tracking.
The following two subsections introduce details of the modality-unified feature extractor and the modality-adaptive box head.
In the last subsection, we introduce the training objectives.

\subsection{Tracking Architecture}
As shown in Figure~\ref{fig:architecture}, UVLTrack can take different modal references as input, including natural language, template, or both.
The template is cropped based on the initial bounding box.
Given the language description $l$, we tokenize the sentence and embed each word with the text embedding layer to obtain text embeddings $\mathbf{E}_l^0\in\mathbb{R}^{N_l\times{C}}$.
$N_l$ is the maximum text length.
Given the template $z\in\mathbb{R}^{3\times{H_z}\times{W_z}}$ and search region (full image for grounding) $x\in\mathbb{R}^{3\times{H_x}\times{W_x}}$, they are split and reshaped into a sequence of flattened 2D patches and then linearly projected into latent space.
Learnable position embeddings are added to the corresponding patch embeddings, obtaining template embeddings $\mathbf{E}_z^0\in \mathbb{R}^{N_z\times C}$ and search region embeddings $\mathbf{E}_x^0\in \mathbb{R}^{N_x\times C}$.
$N_z$ and $N_x$ are the patch number of the template and search region respectively.
We also prepend a language semantic token $\mathbf{T}_l^0\in \mathbb{R}^{1\times C}$ and a visual semantic token $\mathbf{T}_v^0\in \mathbb{R}^{1\times C}$ to text embeddings and image embeddings correspondingly, which are designed to capture the global semantics of different modalities.
After that, text and image embeddings are fed into the modality-unified feature extractor, which is built based on Transformer architecture.
Specifically, we extract language and visual features separately in shallow encoder layers and fuse them in deep encoder layers, which avoids the confusion in low-level feature modeling between different modalities and enables high-level semantics interaction.
Moreover, a multi-modal contrastive loss is proposed to align different modal features into a unified semantic space.
Finally, we feed the enhanced text and image embeddings into the modality-adaptive box head, which can make full use of different modal references to mine ever-changing scenario features from video contexts and locate the target in a contrastive way.
Further, search region embeddings with high-confidence target bounding boxes are saved as video contexts $\mathbf{E}_c^{N+M}$ to help with subsequent target localization.

\begin{figure}[t]
    \centering
    \includegraphics[width=1.0\linewidth]{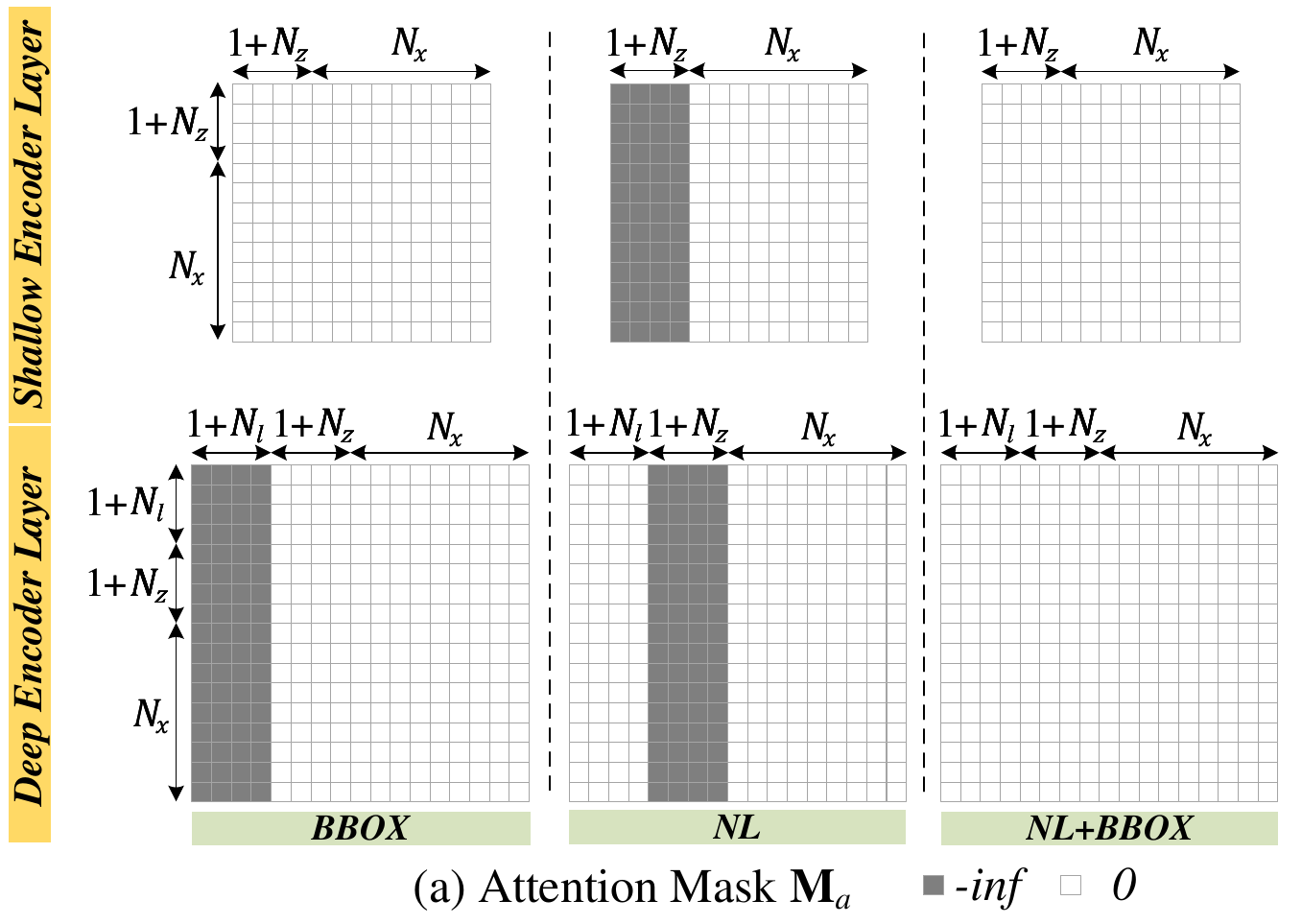}
    \caption{
    The attention mask of task-oriented multi-head attention for different target references.
    }
    \label{fig:attention_mask}
\end{figure}

\subsection{Modality-Unified Feature Extractor}
As shown in Figure~\ref{fig:architecture}, the modality-unified feature extractor is designed based on Transformer architecture, which consists of $N$ shallow encoder layers and $M$ deep encoder layers.
We extract visual and language features separately in shallow encoder layers and fuse them in deep encoder layers, which can avoid the confusion of low-level feature modeling for different modalities and allow high-level semantics interaction for target localization.

For parallel training of different reference inputs, we fill the unavailable (\textbf{NA}) reference embeddings with zeros and propose a task-oriented multi-head attention mechanism (TMHA) to avoid task-irrelevant feature interactions.
We only present the single-head formulas of TMHA below for the sake of simplicity.
Given the input of $i^{th}$ encoder layer $\mathbf{E}^{i-1}$, key ${\mathbf{K}}^i$, query ${\mathbf{Q}}^i$ and value $\mathbf{V}^i$ arise from $\mathbf{E}^{i-1}$ through layer normalization and linear projections.
Then, we filter task-irrelevant feature interactions in attention mechanisms through masking. The output of $i^{th}$ encoder layer can be formulated as,
%
\begin{gather}
    \widehat{\mathbf{E}}^i =
    {\rm Softmax}(\frac{\mathbf{Q}^i(\mathbf{K}^i)^\top}{\sqrt{C}}+\textbf{M}_a)\mathbf{V}^i+\mathbf{E}^{i-1},\\
    \mathbf{E}^{i} = {\rm MLP}({\rm LN}(\widehat{\mathbf{E}}^i)) + \widehat{\mathbf{E}}^i,
\end{gather}
where ${\rm MLP(\cdot)}$ is multi-layer perception, ${\rm LN(\cdot)}$ is layer normalization, $\widehat{\mathbf{E}}^i$ is a intermediate variable, $\mathbf{E}^{i}$ is the output of $i^{th}$ encoder layer. $\textbf{M}_a$ is the attention mask, which is related to the input reference type.
Figure~\ref{fig:attention_mask} shows the details of the attention mask for different target references.

\begin{figure}[t]
    \centering
    \includegraphics[width=1.0\linewidth]{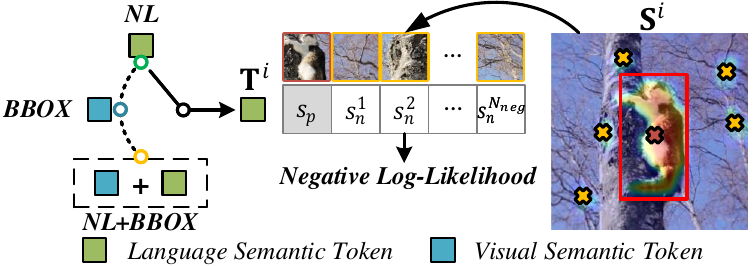}
    \caption{
    The diagram of the multi-modal contrastive loss.
    }
    \label{fig:loss}
\end{figure}

Further, we propose a multi-modal contrastive (MMC) loss to align different modal features into a unified semantic space.
As shown in Figure~\ref{fig:loss}, given the semantic token $\mathbf{T}^i$ of $i^{th}$ encoder layer, we compute the similarity $\mathbf{S}^i=[s^{i,1};s^{i,2},...,s^{i,N_x}]$ between $\mathbf{T}^i$ and search region embeddings $\mathbf{E}_x^i=[f^{i,1};f^{i,2},...,f^{i,N_x}]$.
Formally,
\begin{small}
\begin{gather}
    s^{i,j} = {\rm sim}(\mathbf{T}^i, f^{i,j}) / \tau,
    {\rm sim}(\mathbf{T}^i, f^{i,j}) = \frac{\mathbf{T}^i(f^{i,j})^\top}{||\mathbf{T}^i||_2||f^{i,j}||_2},
\end{gather}
\end{small}

\noindent
where $\tau$ is a temperature parameter, $||\cdot||_2$ means $l_2$ norm.
According to $\mathbf{S}^i$, we select the central score of the target $s_p^i$ as positive sample score and top $N_{neg}$ scores out of the target box $[s_n^{i,k}]_{k=1}^{N_{neg}}$ as negative sample scores.
Finally, the multi-modal contrastive loss can be formulated as follows,
\begin{gather}
    \mathcal{L}_{mmc}^i = -{\rm log}(\frac{e^{s_p^i}}{e^{s_p^i}+\sum_{k=1}^{N_{neg}}e^{s_n^{i,k}}}).
\end{gather}
Different modal features can be aligned to a unified semantic space in a contrastive way, which helps consistent feature learning for different modal references.

\subsection{Modality-Adaptive Box Head}

\begin{figure*}[t]
    \centering
    \includegraphics[width=0.98\linewidth]{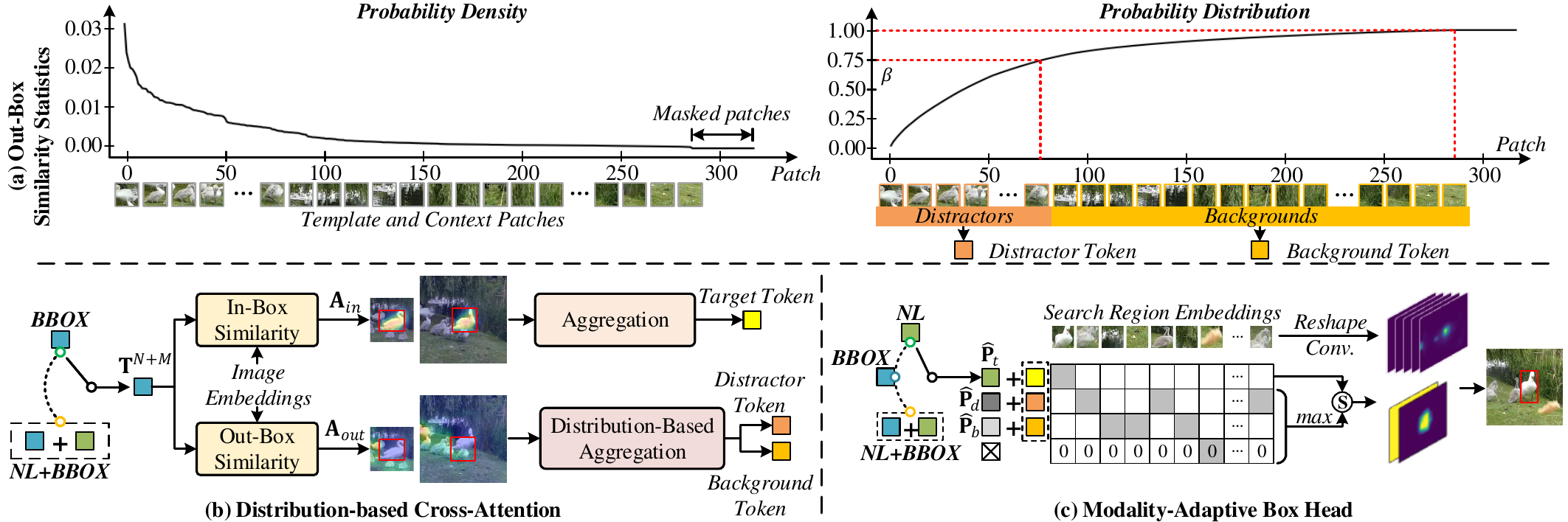}
    \caption{(a) shows the out-box similarity statistics. (b) shows the structure of the distribution-based cross-attention. (c) shows the schematic of the modality-adaptive box head, which can make full use of reference information to discriminate the target.}
    \label{fig:head}
\end{figure*}

Inspired by OSTrack~\cite{OSTrack}, we reshape the reference enhanced embeddings of search region $\textbf{E}_x^{N+M}$ into a 2D feature map and feed it into a three-branch convolutional network to regress a center score map $\hat{\mathbf{C}}\in (0,1)^{\frac{H_x}{p}\times\frac{W_x}{p}}$, an offset map $\hat{\mathbf{O}}\in [0,1)^{2\times\frac{H_x}{p}\times\frac{W_x}{p}}$ and a normalized box size map $\hat{\mathbf{S}}\in (0,1)^{2\times\frac{H_x}{p}\times\frac{W_x}{p}}$, where $p$ is the size of image patches.
However, we find that the center score map is unstable for different modal references, which seriously affects tracking robustness.
%
The underlying reason is that various reference modalities increase the difficulty of static head training.
Thus, we further propose a dynamic head, modality-adaptive box head (MABH), which can make full use of reference information to mine ever-changing scenario features from video context to discriminate the target.

As shown in Figure~\ref{fig:head}(c), we treat the semantic token in the last encoder layer as the target prototype $\widehat{\mathbf{P}}_t=\mathbf{T}^{N+M}$ and introduce learnable distractor prototype $\widehat{\mathbf{P}}_d$ and background prototype $\widehat{\mathbf{P}}_b$ to locate the target in a contrastive way.
For tracking tasks, video contexts can provide rich scenario cues to discriminate the target.
Thus, as shown in Figure~\ref{fig:head}(b), we propose a novel distribution-based attention mechanism to mine features of the target, distractor and background from historic frames.
Given the template and context embeddings $\mathbf{E}_t=[\mathbf{E}_z^{N+M};\mathbf{E}_c^{N+M}]\in\mathbb{R}^{(N_z+N_x)\times{C}}$ and the target masks $\textbf{M}_t=[\textbf{M}_z;\textbf{M}_c]\in\mathbb{R}^{{1}\times(N_z+N_x)}$, we compute in-box similarity $\mathbf{A}_{in}$ and out-box similarity $\mathbf{A}_{out}$ between the target semantic token $\mathbf{T}^{N+M}$ and $\mathbf{E}_t$ to obtain the probability that the patch belongs to the target. Formally,
\begin{gather}
    \mathbf{A}_{in} = {\rm Softmax}(\frac{\mathbf{T}^{N+M}\mathbf{E}_t^\top}{\sqrt{C}}+\textbf{M}_t),\\
    \mathbf{A}_{out} = {\rm Softmax}(\frac{\mathbf{T}^{N+M}\mathbf{E}_t^\top}{\sqrt{C}}+\widetilde{\textbf{M}}_t),
\end{gather}
where $\textbf{M}_t$ is obtained by assigning the position in the target box to 0 and the position out of the target box to $-inf$. 
$\widetilde{\textbf{M}}_t$ is the opposite.
Then, the target token $\mathbf{T}_t$ is obtained by in-box similarity aggregation $\mathbf{T}_t = \mathbf{A}_{in}\mathbf{E}_t$.
Considering that the distractor with a similar appearance to the target is a key factor affecting the tracking robustness~\cite{KeepTrack}, we divide features out of the target box into distractor and background from the perspective of distribution.
Specifically, as shown in Figure~\ref{fig:head}(a), we rank the out-box probabilities $\mathbf{A}_{out}$ in descending order and sum them cumulatively to obtain the probability distribution.
We divide patches by threshold $\beta$ to obtain the distractor mask $\textbf{M}_d$ and background mask $\widetilde{\textbf{M}}_d$.
For distractor mask $\textbf{M}_d$, we assign the patch whose target probability distribution score is lower than $\beta$ to 0 and other positions to $-inf$.
$\widetilde{\textbf{M}}_d$ is the opposite.
Then, the distractor token and background token can be formulated as,
%
\begin{gather}
    \mathbf{T}_d = {\rm Softmax}(\frac{\mathbf{T}^{N+M}\mathbf{E}_t^\top}{\sqrt{C}}+\widetilde{\textbf{M}}_t+\textbf{M}_d)\mathbf{E}_t,\\
    \mathbf{T}_b = {\rm Softmax}(\frac{\mathbf{T}^{N+M}\mathbf{E}_t^\top}{\sqrt{C}}+\widetilde{\textbf{M}}_t+\widetilde{\textbf{M}}_d)\mathbf{E}_t.
\end{gather}
After obtaining $\mathbf{T}_t,\mathbf{T}_d,\mathbf{T}_b$, we add them to scenario prototypes $\widehat{\mathbf{P}}_t,\widehat{\mathbf{P}}_d,\widehat{\mathbf{P}}_b$ to supplement the dynamic scenario information.
Given search region embeddings $\mathbf{E}_x^{N+M}=[f^1;f^2,...,f^{N_x}]$, we compute the corresponding target similarity $\hat{\mathbf{L}}=[{\alpha}^1_t,{\alpha}^2_t,...,{\alpha}^{N_x}_t]$ as follows,
%
\begin{gather}
    {\mathbf{P}}_t = \widehat{\mathbf{P}}_t + \mathbf{T}_t,
    {\mathbf{P}}_d = \widehat{\mathbf{P}}_d + \mathbf{T}_d,
    {\mathbf{P}}_b = \widehat{\mathbf{P}}_b + \mathbf{T}_b,\\
    \hat{\alpha}^i_t = {\rm sim}({f^i, {\mathbf{P}}_t})/\tau, \\
    \hat{\alpha}^i_b = {\rm max}\big({\rm sim}(f^i,{\mathbf{P}}_d)/\tau,{\rm sim}(f^i,{\mathbf{P}}_b)/\tau,0\big),\\
    {\alpha}^i_t = {e^{\hat{\alpha}^i_t}}/{(e^{\hat{\alpha}^i_t}+e^{\hat{\alpha}^i_b})}.
\end{gather}
Here, we append a zero to the background score computation, which avoids unseen objects having relatively high target score $\alpha^i_t$ after the softmax operation.
Finally, given the position $(x_c,y_c)={\rm argmax}_{(x,y)}\hat{\mathbf{C}}(x,y)\hat{\mathbf{L}}(x,y)$, the bounding box of target $\hat{b}=(\hat{x},\hat{y},\hat{w},\hat{h})$ can be formulated as,
\begin{scriptsize}
\begin{gather}
    (\hat{x},\hat{y}) = \Big(\big(x_c+\hat{\mathbf{O}}(0,x_c,y_c)\big)\cdot p,
                             \big(y_c+\hat{\mathbf{O}}(1,x_c,y_c)\big)\cdot p\Big),
\end{gather}
\begin{gather}
    (\hat{w},\hat{h}) = \big(\hat{\mathbf{S}}(0,x_c,y_c)\cdot H_x, \hat{\mathbf{S}}(1,x_c,y_c)\cdot W_x\big).
\end{gather}
\end{scriptsize}

\subsection{Training Objective}
We treat patches in the target box as positive samples and others as negative samples to generate the groundtruth ${\mathbf{L}}$ for target score map $\hat{\mathbf{L}}$.
Then, the binary cross-entropy loss is adopted for the target score map constraint.
Formally,
\begin{gather}
    \mathcal{L}_{tgt} = \mathcal{L}_{bce}(\hat{\mathbf{L}}, {\mathbf{L}})
\end{gather}
The training objectives of center score map $\mathcal{L}_{cls}$ and bounding box $\mathcal{L}_{box}=\lambda_1\mathcal{L}_1+\lambda_{giou}\mathcal{L}_{giou}$ are consistent with OSTrack~\cite{OSTrack}.
In summary, the overall objective function can be formulated as,
\begin{gather}
    \mathcal{L} = \mathcal{L}_{tgt}+\mathcal{L}_{cls}+\mathcal{L}_{box}+\lambda_{mmc}\sum_{i=1}^{N+M}\mathcal{L}_{mmc}^{i}.
\end{gather}

\section{Experiment}

\begin{table*}[!t]
\begin{center}
\resizebox{0.93\linewidth}{!}{
\begin{tabular}{c|cc|ccc|cc|cc|cc}
\hline
\multirow{2}{*}{Method}	&\multicolumn{2}{c|}{TNL2K}
&\multicolumn{3}{c|}{AVisT}
&\multicolumn{2}{c|}{LaSOT}
&\multicolumn{2}{c|}{LaSOT$_{ext}$}
&\multicolumn{2}{c}{TrackingNet} \\
\cline{2-12}
&AUC	&P &AUC	&OP$_{0.5}$	&OP$_{0.75}$ &AUC	&P	&AUC	&P  &AUC	&P  \\ 
\hline
\multicolumn{12}{c}{\textbf{Performance-oriented Variants}}\\
\hline
\rowcolor{mygray}
UVLTrack-L 		                        &\textbf{{64.8}}
                                        &\textbf{{68.8}}
                                        &\textbf{{57.8}}
                                        &\textbf{{67.9}}
                                        &\textbf{{48.7}}
                                        &\textbf{{71.3}}
                                        &\textbf{{78.3}}
                                        &\textbf{{51.2}}
                                        &\textbf{{59.0}}
                                        &\textbf{{84.1}}
                                        &82.9 \\
OSTrack-384~\cite{OSTrack}              &\underline{{55.9}}
                                        &-
                                        &-
                                        &-
                                        &-
                                        &\underline{{71.1}}
                                        &\underline{{77.6}}
                                        &\underline{{50.5}}
                                        &\underline{{57.6}}
                                        &\underline{{83.9}}
                                        &\textbf{{83.2}}\\
MixFormer-L~\cite{cui2022mixformer}     &-      &-    &\underline{{56.0}}   &\underline{{65.9}}   &\underline{{46.3}}  &70.1   &76.3   &-      &-      &\underline{{83.9}}     &\underline{{83.1}}       \\ 
SimTrack-L/14~\cite{SimTrack}           &55.6   &\underline{{55.7}}   &-      &-      &-    &70.5    &-      &-          &-      &83.4   &-         \\ 
\hline
\multicolumn{12}{c}{\textbf{Basic Variants}}\\
\hline
\rowcolor{mygray}
UVLTrack-B 		                        &\textbf{{62.7}}
                                        &\textbf{{65.4}}
                                        &\textbf{{56.5}}
                                        &\textbf{{66.0}}
                                        &\textbf{{45.1}}
                                        &\textbf{{69.4}}
                                        &\underline{{74.9}}
                                        &\textbf{{49.2}}
                                        &\textbf{{55.8}}
                                        &\textbf{{83.4}}
                                        &\textbf{{82.1}} \\
OSTrack-256~\cite{OSTrack}              &54.3   &-      &-      &-      &-    &69.1  &\textbf{{75.2}}   &47.4   &53.3   &\underline{{83.1}}   &\underline{{82.0}}       \\
MixFormer-22k~\cite{cui2022mixformer}       &-      &-      &\underline{{53.7}}   &\underline{{63.0}}   &\underline{{43.0}}   &69.2   &74.7   &-      &-      &\underline{{83.1}}    &81.6       \\
SimTrack-B/16~\cite{SimTrack}           &\underline{{54.8}}   &\underline{{53.8}}   &-      &-      &-    &\underline{{69.3}}   &-      &-      &-      &82.3   &-          \\
AiATrack~\cite{AiATrack}                &-      &-      &-      &-      &-   &69.0   &73.8   &\underline{{47.7}}     &\underline{{55.4}}   &82.7   &80.4       \\
STARK~\cite{yan2021learning}		    &-      &-	    &51.1   &59.2   &39.1   &66.4	&71.2   &-      &-      &81.3	&78.1       \\
TransT~\cite{TransT}		            &50.7   &51.7   &49.0   &56.4   &37.2   &64.9	&73.8   &-      &-	    &81.4	&80.3       \\
TrDiMP~\cite{wang2021transformer}       &-      &-      &48.1   &55.3   &33.8   &63.9   &61.4   &-      &-      &78.4   &73.1       \\
STMTrack~\cite{fu2021stmtrack}          &-      &-      &-      &-      &-      &60.6   &63.3   &-      &-      &80.3   &76.7       \\
PrDiMP~\cite{PrDiMP}                    &47.0   &45.9   &43.3   &48.0   &28.7   &59.8	&60.8   &-      &-	    &75.8	&70.4       \\
SiamFC++~\cite{SiamFC++}		        &38.6   &36.9   &-      &-      &-      &54.4	&54.7   &-      &-  	&75.4	&70.5	    \\
Ocean~\cite{Ocean}		                &38.4   &37.7   &38.9   &43.6   &20.5   &56.0	&56.6	&-      &-      &-	    &-	        \\
DiMP~\cite{DiMP}    	                &44.7   &43.4   &41.9   &45.7   &26.0   &56.9	&56.7	&39.2   &45.1   &74.0	&68.7       \\
SiamRPN++~\cite{SiamRPNplusplus}	    &41.3   &41.2   &39.0   &43.5   &21.2   &49.6	&49.1   &34.0   &39.6	&73.3	&69.4       \\
SiamFC\cite{SiameseFC}                  &29.5   &28.6   &-      &-      &-      &33.6   &33.9   &23.0   &26.9   &57.1   &53.3       \\
\hline
\end{tabular}
}
\caption{Comparison with state-of-the-art visual trackers on TNL2K, AVisT, LaSOT, LaSOT$_{ext}$ and TrackingNet. The best two results are shown in bold and underline}
\label{tab:mainresults}
\end{center}
\end{table*}

\begin{table*}[t]
\begin{center}
\resizebox{0.9\linewidth}{!}{
\begin{tabular}{c|cccccc|cc}
\hline
& Ocean
& DiMP-50
& PrDiMP-50
& TransT
& OSTrack-256 & OSTrack-384
& UVLTrack-B  & UVLTrack-L \\
\hline
NFS      &49.4    &61.8
                    &63.5    &65.3
                    &64.7    &\underline{{66.5}}
                    &65.9
                    &\textbf{{67.6}}\\
UAV123   &57.4    &64.3
                    & 68.0   &68.1
                    &68.3    &\underline{{70.7}}
                    &69.3
                    &\textbf{{71.0}}\\
\hline
\end{tabular}
}
\caption{Comparison with state-of-the-art trackers on NFS, UAV123 datasets in terms of overall AUC score. The best two results are shown in bold and underline}
\label{tab:smallresults}
\end{center}
\end{table*}

Our tracker is implemented using Python 3.8.13 and Pytorch 1.10.1. The experiments are conducted on a server with eight 24GB NVIDIA RTX 3090 GPUs. Visualization and qualitative results are present in \textbf{Supplementary Materials}.
\subsection{Implementation Details}

\textbf{Network Details.}
We crop the template and search region by $2^2$ and $4^2$ times the target bounding box area and resize them to 128$\times$128 and 256$\times$256 respectively.
%
%
The test image for first frame grounding is scaled such that its long edge is 256.
Image patch size $p$=16.
%
For language, the max length of the sentence $N_l$ is set to 40.
To demonstrate the scalability of UVLTrack, we present two variants, termed UVLTrack-B and UVLTrack-L.
The number of encoder layers is set to $N$=6,$M$=6 for UVLTrack-B and $N$=12,$M$=12 for UVLTrack-L.
The language branch in shallow encoder layers is initialized with uncased parameters of BERT~\cite{BERT}.
Other parameters in the modality-unified feature extractor are initialized with ViT parameters pretrained by MAE~\cite{he2022masked}.
The modality-adaptive box head is initialized with Xavier init~\cite{Xavier}.

\noindent
\textbf{Training Details.}
We train our model on the training splits of LaSOT~\cite{LaSOT}, GOT-10k~\cite{GOT10K}, COCO2017~\cite{COCO}, TrackingNet~\cite{trackingnet}, TNL2K~\cite{TNL2K}, OTB99~\cite{nltrack} and RefCOCOg-google~\cite{RefCOCOg}.
Common data augmentation is used for model training, such as translation, horizontal flip, and color jittering.
Due to the flexibility of our framework, different modal references can be trained jointly, which provides a neat training pipeline.
The loss weights are set to $\lambda_{giou}$=2.0, $\lambda_1$=5.0, $\lambda_{mmc}$=0.1.
%

\subsection{State-of-the-art Comparisons}

\noindent
\textbf{Visual Tracking.}
We evaluate our trackers on seven visual tracking benchmarks, including TNL2K~\cite{TNL2K}, AVisT~\cite{AVisT}, LaSOT~\cite{LaSOT}, LaSOT$_{ext}$~\cite{LaSOT_ext}, TrackingNet~\cite{trackingnet}, NFS~\cite{NFS} and UAV123~\cite{UAV}, which are commonly used for visual tracker evaluation.
The Area Under the Curve (AUC) of the success plot is the main metric to rank trackers.
As shown in Table~\ref{tab:mainresults} and \ref{tab:smallresults}, our UVLTrack-B outperforms the best visual tracking counterparts.
Further, UVLTrack-L achieves new state-of-the-art performance on all seven visual tracking benchmarks.
These results demonstrate the effectiveness of UVLTrack under the bounding box reference setting (\textbf{BBOX}).

\begin{table*}[!t]
\begin{center}
\resizebox{0.7\linewidth}{!}{
\begin{tabular}{c|ccc|ccc|ccc}
\hline
\multirow{2}{*}{Method}
&\multicolumn{3}{c|}{RefCOCO}
&\multicolumn{3}{c|}{RefCOCO+}
&\multicolumn{3}{c }{RefCOCOg} \\
\cline{2-10}
&\textit{val} &\textit{testA} &\textit{testB} &\textit{val} &\textit{testA} &\textit{testB} &\textit{val-g} &\textit{val-u} &\textit{test-u}   \\
\hline
\rowcolor{mygray}
UVLTrack*                        &\textbf{{85.47}}     &\textbf{{87.56}}     &\underline{{81.73}}     &\textbf{{74.60}}     &\textbf{{79.70}}    &\textbf{{65.64}}    &\textbf{{73.86}}    &\underline{{75.94}}    &\textbf{{74.86}}        \\
VLTVG  &\underline{{84.77}}   &\underline{{87.24}}   &80.49   &\underline{{74.19}}   &\underline{{78.93}}  &\underline{{65.17}}  &\underline{{72.98}}  &\textbf{{76.04}}    &74.18      \\
SeqTR       &83.72   &86.51   &81.24   &71.45   &76.26  &64.88  &71.50        
&74.86    &\underline{{74.21}}      \\
QRNet     &84.01   &85.85   &\textbf{{82.34}}   &72.94   &76.17  &63.81  &71.89
&73.03    &72.52      \\
TransVG  &81.02   &82.72   &78.35   &64.82   &70.70  &56.94  &67.02  &68.67    &67.73      \\
Ref-NMS      &80.70   &84.00   &76.04   &68.25   &73.68  &59.42  &-
&70.55    &70.62                      \\
LBYL-Net    &79.67 &82.91  &74.15   &68.64   &73.38  &59.49  &62.70
&-    &-      \\
ReSC-Large &77.63 &80.45  &72.30   &63.59   &68.36  &56.81  &63.12
&67.30    &67.20\\
NMTree   &76.41   &81.21   &70.09   &66.46   &72.02  &57.52  &64.62  
&65.87    &66.44      \\
\hline
\end{tabular}
}
\caption{Comparison with state-of-the-art grounding methods on RefCOCO, RefCOCO+, RefCOCOg datasets.}
\label{tab:grresults}
\end{center}
\end{table*}

\noindent
\textbf{Discussion.}
We evaluate recent vision-language trackers initialized by the target bounding box on LaSOT.
JointNLT achieves 54.5\% AUC and VLTTT achieves 53.4\% AUC.
These vision-language trackers show limited performance without natural language.
This is because these trackers ignore the semantic gap between different modalities and tend to rely on semantic information in language.
Differently, UVLTrack aligns vision and language into a unified semantic space, providing a unified tracking framework, which delivers superior performance for all three reference settings.

\noindent
\textbf{Vision-Language Tracking.}
We further evaluate our UVLTrack on vision-language tracking benchmarks, including TNL2K~\cite{TNL2K}, LaSOT~\cite{LaSOT} and OTB99~\cite{nltrack} and compare with the latest trackers, including JointNLT~\cite{JointNLT}, CTRNLT~\cite{TSN}, TNL2K~\cite{TNL2K}, GTI~\cite{GTI}, TNLS~\cite{nltrack}, VLTTT~\cite{VLT}, SNLT~\cite{SNLT}.
As shown in Table~\ref{tab:vlresults}, when specifying the target by natural language (\textbf{NL}), UVLTrack-L surpasses the previous best tracker JointNLT with a large margin on three benchmarks.
Also, our UVLTrack-L achieves the best performance on TNL2K and LaSOT when initializing the tracker with both natural language and the bounding box (\textbf{NL+BBOX}).
These results demonstrate the superiority of UVLTrack for vision-language tracking.

\begin{table}[!t]
\begin{center}
\resizebox{0.95\linewidth}{!}{
\begin{tabular}{c|cc|cc|cc}
\hline
\multirow{2}{*}{Method}
&\multicolumn{2}{c|}{TNL2K}
&\multicolumn{2}{c|}{LaSOT}
&\multicolumn{2}{c }{OTB99} \\
\cline{2-7}
&AUC   &P   &AUC   &P   &AUC   &P   \\
\hline
\multicolumn{7}{c}{\textbf{NL}}\\
\hline
\rowcolor{mygray}
UVLTrack-L                      &\textbf{{58.2}}     &\textbf{{60.9}}    &\textbf{{59.6}}     &\textbf{{63.9}}          &\textbf{{63.5}}    &\textbf{{83.2}}        \\
\rowcolor{mygray}
UVLTrack-B                      &\underline{{55.7}}    &\underline{{57.2}}    &\underline{{57.2}}    &\underline{{61.0}}        &\underline{{60.1}}   &\underline{{79.1}}       \\
JointNLT	    &54.6    &55.0     &56.9    &59.3        &59.2   &77.6	    \\
CTRNLT    	        &14.0    &9.0      &52.0    &51.0        &53.0   &72.0	    \\
TNL2K-1		    &11.0    &6.0      &51.0    &49.0        &19.0   &24.0       \\
GTI                  &-       &-        &47.8    &47.6        &58.1   &73.2	    \\
TNLS-II          &-       &-        &-       &-           &25.0   &29.0       \\
\hline
\multicolumn{7}{c}{\textbf{NL+BBOX}}\\
\hline
\rowcolor{mygray}
UVLTrack-L                      &\textbf{{64.9}}     &\textbf{{69.3}}       &\textbf{{71.4}}     &\textbf{{78.7}}          &\underline{{71.1}}    &\underline{{92.0}}        \\
\rowcolor{mygray}
UVLTrack-B                      &\underline{{63.1}}    &\underline{{66.7}}        &\underline{{69.4}}    &\underline{{75.9}}        &69.3   &89.9       \\
JointNLT	    &56.9    &58.1    &60.4    &63.6        &65.3   &85.6	    \\
VLTTT    	        &53.1    &53.3    &67.3    &72.1        &\textbf{{76.4}}   &\textbf{{93.1}}	    \\
SNLT               &27.6    &41.9    &54.0    &57.6        &66.6   &80.4	    \\
TNL2K-2		    &42.0    &42.0    &51.0    &55.0        &68.0   &88.0       \\
TNLS-III         &-       &-       &-       &-           &55.0   &72.0       \\
\hline
\end{tabular}
}
\caption{Comparison with state-of-the-art vision-language trackers on LaSOT, TNL2K and OTB99 datasets.}
\label{tab:vlresults}
\end{center}
\end{table}

\noindent
\textbf{Efficiency.}
UVLTrack-B runs at 58 FPS for visual tracking and 57 FPS for vision-language tracking.
UVLTrack-L runs at 28 FPS for visual tracking and 27 FPS for vision-language tracking.
Compared with JointNLT (39 FPS), UVLTrack-B achieves better performance with \textbf{1.46$\times$} speed.

\noindent
\textbf{Visual Grounding.}
Like VLTVG, we retrain UVLTrack-B on train sets of RefCOCO~\cite{RefCOCO}, RefCOCO+~\cite{RefCOCO} and RefCOCOg~\cite{RefCOCOg} separately, and report the Top-1 accuracy on corresponding test sets in Table~\ref{tab:grresults}.
We compared UVLTrack with the latest grounding methods, including VLTVG~\cite{yang2022improving}, SeqTR~\cite{zhu2022seqtr}, QRNet~\cite{ye2022shifting}, TransVG~\cite{deng2021transvg}, Ref-NMS~\cite{chen2021ref}, LBYL-Net~\cite{huang2021look}, ReSC-Large~\cite{yang2020improving}, NMTree~\cite{liu2019learning}
The test image is scaled such that its long edge is 384.
Compared to visual grounding methods, our UVLTrack achieves the best performance on seven test sets, demonstrating the generalization ability of our framework.

\subsection{Ablation Study}
The following experiments use UVLTrack-B as the base model.
The baseline is UVLTrack without MMCLoss constraint and using the static anchor-free head for localization.

\begin{figure}[t]
    \centering
    \includegraphics[width=1.0\linewidth]{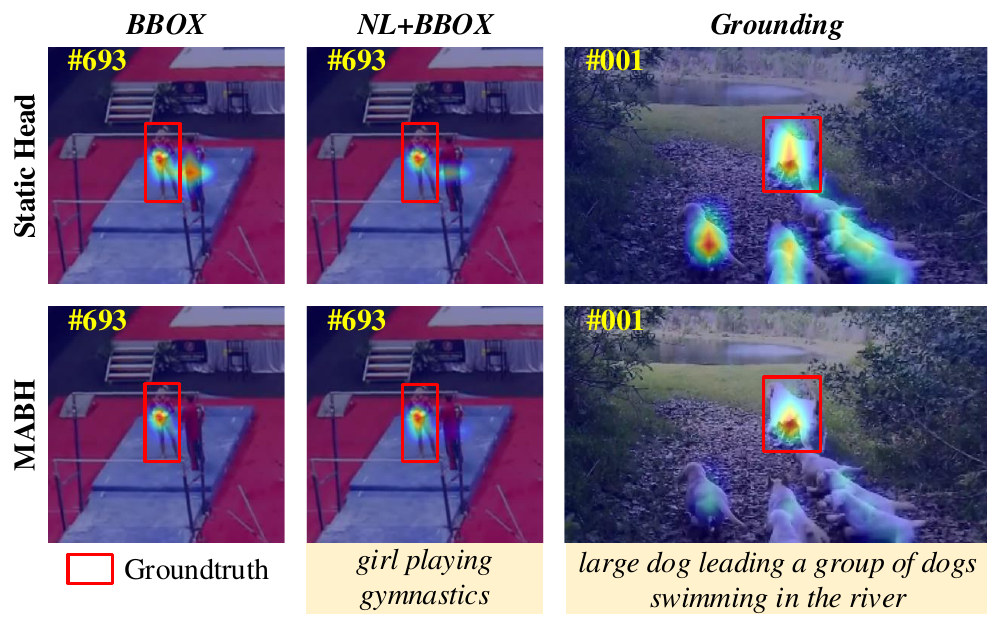}
    \caption{
    Visualization of target localization results.
    }
    \label{fig:visualization}
\end{figure}

\begin{table}[!t]
\begin{center}
\resizebox{0.95\linewidth}{!}{
\begin{tabular}{c|cc|cc|cc}
\hline
\multirow{3}{*}{Method}
&\multicolumn{6}{c}{TNL2K}\\
\cline{2-7}
&\multicolumn{2}{c|}{BBOX}
&\multicolumn{2}{c|}{NL}
&\multicolumn{2}{c}{NL+BBOX} \\
\cline{2-7}
&AUC &P &AUC &P &AUC &P   \\
\hline
baseline            &59.4     &59.9     &51.6    &51.8    &59.7   &60.6    \\
+MMCLoss            &60.6    &62.8    &53.8    &54.7    &62.0   &64.9    \\
\rowcolor{mygray}
+MABH            &\textbf{62.7}     &\textbf{65.4}     &\textbf{55.7}     &\textbf{57.2}     &\textbf{63.1}    &\textbf{66.7}     \\
\hline
\end{tabular}
}
\caption{Analysis of different components in UVLTrack.}
\label{tab:components}
\end{center}
\end{table}

\begin{table}[t]
\begin{center}
\resizebox{0.9\linewidth}{!}{
\begin{tabular}{c|c|cc|cc|cc}
\hline
\multirow{3}{*}{$N$}
&\multirow{3}{*}{$M$}
&\multicolumn{6}{c}{TNL2K}\\
\cline{3-8}
&&\multicolumn{2}{c|}{BBOX}
&\multicolumn{2}{c|}{NL}
&\multicolumn{2}{c}{NL+BBOX} \\
\cline{3-8}
&&AUC &P &AUC &P &AUC &P   \\
\hline
0 & 12 &61.3     &63.8     &55.1     &56.1     &62.3    &65.1     \\
3 & 9  &61.6     &64.2     &\textbf{55.9}     &\textbf{57.3}     &63.0    &66.6     \\
\rowcolor{mygray}
6 & 6  &\textbf{62.7}     &\textbf{65.4}     &55.7     &57.2     &\textbf{63.1}    &\textbf{66.7}     \\
9 & 3  &62.6     &65.3     &54.6     &55.4     &62.6    &65.8     \\
11 & 1 &62.5     &65.1     &46.2     &42.3     &61.9    &64.6     \\
\hline
\end{tabular}
}
\caption{Analysis of the modality-unified feature extractor.}
\label{tab:backbone}
\end{center}
\end{table}

\noindent
\textbf{Effectiveness of the Different Components.}
Table~\ref{tab:components} shows the performance of UVLTrack with different components.
MMCLoss brings 1.2\%, 2.2\% and 2.3\% AUC gains for BBOX, NL and BBOX+NL reference settings respectively.
This is because MMCLoss can align different modal features into a unified semantic space, which enables consistent feature learning for different reference modalities.
The modality-adaptive box head (MABH) brings 2.1\%, 1.9\% and 1.1\% AUC gains for BBOX, NL and BBOX+NL reference settings respectively.
As shown in Figure~\ref{fig:visualization}, the static anchor free head shows unstable target localization results.
The reason is that various reference modalities increase the difficulty of static head training, which leads to compromised results.
Differently, we design a dynamic head (MABH), which can make full use of reference information to locate the target in a contrastive way, improving tracking performance across all reference settings.

\noindent
\textbf{Analysis of the Modality-Unified Feature Extractor.}
As shown in Table~\ref{tab:backbone}, more separate layers (larger $N$) are beneficial for visual tracking and more fusion layers (larger $M$) are beneficial for vision-language tracking.
However, when we fuse visual and language features in all encoder layers, the performance is suboptimal for vision-language tracking.
This is because early fusion breaks the low-level feature modeling for different modalities.
Thus, we set $N$=6 and $M$=6 to balance the performance for all reference settings.

\noindent
\textbf{Analysis of the Multi-Modal Contrastive Loss.}
We study different ways to obtain the positive sample and the negative sample.
As shown in Table~\ref{tab:mmcloss}, the best results are achieved when we sample the central score of the target as the positive sample and the top 9 scores out of the target box as negative samples.
The underlying reason is that the central feature of the target contains no backgrounds, which is more reliable to express the target.
Further, more hard-negative samples can improve the discriminability of the semantic token and align different modal features into a compact semantic space.

\noindent
\textbf{Analysis of the Distractor Threshold.}
As shown in Table~\ref{tab:distractor_thres}, the performance of UVLTrack is insensitive to threshold $\beta$ over a wide range.
However, the performance drops overtly if we aggregate all background features into one token ($\beta$=0).
This is because the distractor feature is vital to discriminate the target in complex scenarios.
If we aggregate all background features together, the distractor features will be smoothed by other background features, which is not conducive for the tracker to distinguish distractors.

\noindent
\textbf{Analysis of the Training Strategy.}
As shown in Table~\ref{tab:training_strategy}, the ratio of different references (BBOX:NL:Both) has little effect on the performance of UVLTrack.
Moreover, we try to initialize all parameters in the backbone with ViT parameters pretrained by MAE, which reduces overall performance.
This is because pretrained BERT parameters bring initial language modeling capabilities to the tracker to understand natural language.

\begin{table}[!t]
\begin{center}
\resizebox{\linewidth}{!}{
\begin{tabular}{c|c|c|cc|cc|cc}
\hline
\makebox[11pt][c]{\multirow{3}{*}{pos.}}
&\multirow{3}{*}{neg.}
&\makebox[14pt][c]{\multirow{3}{*}{$N_{neg}$}}
&\multicolumn{6}{c}{TNL2K}\\
\cline{4-9}
&&&\multicolumn{2}{c|}{BBOX}
&\multicolumn{2}{c|}{NL}
&\multicolumn{2}{c}{NL+BBOX} \\
\cline{4-9}
&&&\makebox[11pt][c]{AUC} &P &\makebox[11pt][c]{AUC} &P &\makebox[11pt][c]{AUC} &P   \\
\hline
\makebox[11pt][c]{$avg$} & \makebox[15pt][c]{$rand$} & 9  &61.4    &63.8     &53.6     &54.0     &61.5    &64.2     \\
$ctr$ & \makebox[15pt][c]{$rand$} & 9  &61.8    &64.5     &54.9     &55.9     &62.1    &65.4     \\
\makebox[11pt][c]{$avg$} & $top$  & 9  &62.1    &64.6     &55.2     &56.3     &62.3    &65.5     \\
\rowcolor{mygray}
$ctr$ & $top$  & 9  &\textbf{62.7}    &\textbf{65.4}     &\textbf{55.7}     &\textbf{57.2}     &\textbf{63.1}    &\textbf{66.7}     \\
$ctr$ & $top$  & 1  &61.6    &64.1     &54.5     &55.4     &61.8    &65.1     \\
$ctr$ & $top$  & 5  &62.3    &64.9     &55.4     &56.7     &62.6    &66.0     \\
$ctr$ & $top$  & 13 &62.6    &\textbf{65.4}     &55.6     &57.0     &\textbf{63.1}    &66.6     \\
\hline
\end{tabular}
}
\caption{Analysis of the multi-modal contrastive loss design.}
\label{tab:mmcloss}
\end{center}
\end{table}

\begin{table}[!t]
\begin{center}
\resizebox{0.86\linewidth}{!}{
\begin{tabular}{c|cc|cc|cc}
\hline
\multirow{3}{*}{$\beta$}
&\multicolumn{6}{c}{TNL2K}\\
\cline{2-7}
&\multicolumn{2}{c|}{BBOX}
&\multicolumn{2}{c|}{NL}
&\multicolumn{2}{c}{NL+BBOX} \\
\cline{2-7}
&AUC &P &AUC &P &AUC &P   \\
\hline
0.00               &61.3    &63.8     &54.3     &55.3     &61.8    &64.7     \\
0.25               &62.1    &64.7     &55.2     &56.5     &62.4    &65.8     \\
0.50               &62.5    &65.3     &55.4     &56.8     &62.8    &66.3     \\
\rowcolor{mygray}
0.75               &\textbf{62.7}    &\textbf{65.4}     &\textbf{55.7}     &\textbf{57.2}     &\textbf{63.1}    &\textbf{66.7}     \\
0.85               &62.6    &65.3     &55.3     &56.6     &62.7    &66.2     \\
\hline
\end{tabular}
}
\caption{Analysis of the distractor threshold $\beta$ in the distribution-based cross-attention mechanism.}
\label{tab:distractor_thres}
\end{center}
\end{table}

\begin{table}[!t]
\begin{center}
\resizebox{1.0\linewidth}{!}{
\begin{tabular}{c|c|cc|cc|cc}
\hline
\multirow{3}{*}{Pretrain}
&\multirow{3}{*}{Ratio}
&\multicolumn{6}{c}{TNL2K}\\
\cline{3-8}
&&\multicolumn{2}{c|}{BBOX}
&\multicolumn{2}{c|}{NL}
&\multicolumn{2}{c}{NL+BBOX} \\
\cline{3-8}
&&\makebox[1pt][c]{AUC} &P &\makebox[1pt][c]{AUC} &P &\makebox[1pt][c]{AUC} &P   \\
\hline
\makebox[45pt][c]{BERT+MAE} &3:1:3               &62.3     &65.0     &55.6     &57.0     &62.8    &66.5     \\
\rowcolor{mygray}
\makebox[45pt][c]{BERT+MAE} &4:1:4               &62.7     &65.4     &\textbf{55.7}     &\textbf{57.2}     &63.1    &66.7     \\
\makebox[45pt][c]{BERT+MAE} &5:1:5               &\textbf{62.8}     &\textbf{65.6}     &55.2     &56.5     &\textbf{63.2}    &\textbf{67.0}     \\
MAE      &4:1:4               &62.1     &64.2     &53.8     &54.5     &61.9    &65.1     \\
\hline
\end{tabular}
}
\caption{Analysis of the training strategy.}
\label{tab:training_strategy}
\end{center}
\end{table}


\section{Conclusion}
%
In this work, we propose a novel unified tracker (UVLTrack) for visual and vision-language tracking, which can simultaneously cope with three types of target reference (BBOX, NL, NL+BBOX). 
Specifically, we design a modality-unified feature extractor for joint visual and language feature learning, and propose a multi-modal contrastive loss to align different modal features into a unified semantic space.
Further, a modality-adaptive box head is proposed to localize the target dynamically with scenario features, enabling robust performance for different reference settings.
UVLTrack achieves promising results on seven visual tracking, three vision-language tracking, three visual grounding datasets.

\section{Acknowledgments}
This work was supported by National Natural Science Foundation of China (62306294, 62071122, 62121002) and Youth Innovation Promotion Association CAS (2018166).

\bibliography{aaai24}

\end{document}